# Visual-UWB Navigation System for Unknown Environments

Qin Shi, Xiaowei Cui*, Wei Li, *Tsinghua University, China*
Yu Xia, *North Information Control Research Academy Group Co., Ltd, China*
Mingquan Lu, *Tsinghua University, China*
*Correspondence Email: cwx2005@tsinghua.edu.cn

**BIOGRAPHY (IES)**

**Qin Shi** is a Ph.D. candidate in the Department of Electronic Engineering at Tsinghua University, China. He received his B.S. degree from Tsinghua University in 2015. His current research focuses on indoor localization, sensor fusion and simultaneous localization and mapping.

**Xiaowei Cui** is an associate professor at the Department of Electronic Engineering, Tsinghua University, China. He is a member of the Expert Group of China BeiDou Navigation Satellite System. His research interests include robust GNSS signal processing, multipath mitigation techniques, and high-precision positioning.

**Wei Li** received her B.S. degree in Electronic Engineering from Tsinghua University in 2014. Currently she is pursuing her Ph.D. degree at the Department of Electronic Engineering, Tsinghua University, China. Her major research focuses on precise positioning algorithms and the realization of integrated navigation system.

**Yu Xia** is an engineer of North Information Control Research Academy Group Co., Ltd, China. He received his B.E. degree from Southeast University in 2009 and M.E. degree from Zhejiang University in 2012. His current research focuses on application of GNSS high precision positioning.

**Mingquan Lu** is a professor of the Department of Electronic Engineering, Tsinghua University, China. He is the director of Tsinghua Position, Navigation and Timing Center, and a member of the Expert Group of China BeiDou Navigation Satellite System. His current research interests include GNSS signal design and analysis, GNSS signal processing and receiver development, and GNSS system modeling and simulation.

**ABSTRACT**

Navigation applications relying on the Global Navigation Satellite System (GNSS) are limited in indoor environments and GNSS-denied outdoor terrains such as dense urban or forests. In this paper, we present a novel accurate, robust and low-cost GNSS-independent navigation system, which is composed of a monocular camera and Ultra-wideband (UWB) transceivers. Visual techniques have gained excellent results when computing the incremental motion of the sensor, and UWB methods have proved to provide promising localization accuracy due to the high time resolution of the UWB ranging signals. However, the monocular visual techniques with scale ambiguity are not suitable for applications requiring metric results, and UWB methods assume that the positions of the UWB transceiver anchor are pre-calibrated and known, thus precluding their application in unknown and challenging environments. To this end, we advocate leveraging the monocular camera and UWB to create a map of visual features and UWB anchors. We propose a visual-UWB Simultaneous Localization and Mapping (SLAM) algorithm which tightly combines visual and UWB measurements to form a joint non-linear optimization problem on Lie-Manifold. The 6 Degrees of Freedom (DoF) state of the vehicles and the map are estimated by minimizing the UWB ranging errors and landmark reprojection errors. Our navigation system starts with an exploratory task which performs the real-time visual-UWB SLAM to obtain the global map, then the navigation task by reusing this global map. The tasks can be performed by different vehicles in terms of equipped sensors and payload capability in a heterogeneous team. We validate our system on the public datasets, achieving typical centimeter accuracy and 0.1% scale error.

**INTRODUCTION**

High-precision autonomous navigation of an unmanned ground vehicle (UGV) or an unmanned aerial vehicle (UAV) has long been an important task in Global Navigation Satellite System (GNSS) denied environments, such as urban areas or indoor environments. Generally, a map representing the environment is readily available, thus enabling the vehicle to localize itself against the map. However, for unknown environments which have never been explored, the corresponding map is not available, and the position of the vehicle is also unknown, thus limiting the navigation applications. For navigation purpose, a map is always built once and for all. Building a map while exploring in unknown environments is often referred to as the Simultaneous Localization and Mapping (SLAM) problem by the community [1].

Visual navigation systems utilizing only a monocular camera, have become popular due to their light-weight, low-cost and rich presentation of the environments [2-4]. The visual navigation map represents the construction of the environment by consisting of a set of point features located in world reference frame. To build the map in advance, a visual SLAM algorithm is performed. The 6 Degrees of Freedom (DoF) poses of a vehicle equipped with a single camera and the sparse geometrical reconstruction of the environment are obtained by bundle adjustment (BA) [5], given that the visual feature matches and good initial guesses are provided. After the algorithm is terminated, a fully optimized map is obtained. Then the navigation of the vehicle is performed by finding the corresponding observations of the map features. However, reusing the map is computationally expensive, often a GPU acceleration method is favored, thus restricting the application implementation to a resource-limited vehicle. Furthermore, monocular SLAM suffers from the well-known scale ambiguity, thus raising the problem of the estimation of vehicle's absolute velocity and position.

Recently, we have seen the trend of utilizing Ultra-wideband (UWB) as a navigation system [6-9]. With the development of UWB hardware (such as Decawave's DW1000 chip), UWB navigation system has become low-cost, low-power and high-precision. Typically, a UWB transceiver attached to a vehicle, namely a tag is to be located. Several transceivers, namely anchors are deployed at known locations communicating with anchors. The tag's position is estimated using the metric measurements from UWB signals such as time of flight (TOF) measurements. UWB navigation systems require little computational complexity compared with visual systems. And UWB technology is often used to determine the vehicle's tridimensional position (3 DoF), which is sufficient for most navigation purpose applications. However, as a navigation system, the UWB-only method can rarely enable the vehicle to avoid collisions, due to the fact that a comprehensive map of the environment cannot be obtained. As a passive localization method, the positions of the fixed anchor should be pre-calibrated and known. However, in unknown environments, especially in areas unreachable for mankind, the anchor positions as a prior is not available, and the process of establishing the constellation of the anchors cannot be conducted. This motivates a UWB SLAM method, in which, the mapping indicates the positioning of the anchors, and the anchors are automatically deployed during SLAM progress.

In this work, we advocate utilizing the complementary integration of UWB and monocular camera for a Visual-UWB navigation system in an unknown environment without UWB anchor positions as a prior.

Our system firstly performs an exploratory task as shown in Fig.1: a vehicle equipped with a UWB tag and monocular camera, called as a pioneer, explores the unknown environment. While exploring, it drops and settles UWB anchors to constitute a UWB constellation. In our system, the map is augmented by consisting of a set of point features in world reference and additional dropped anchors' positions. A tightly-coupled visual-UWB SLAM algorithm was proposed to build the map while tracking the vehicle. Due to UWB ranging measurements, the built map is with a metric scale, and robust to challenging visual problems such as motion blur and low light. This fusion strategy yields precise positioning of the dropped anchors and rich representation of the environment. Notably, this exploratory task is conducted without human intervention.

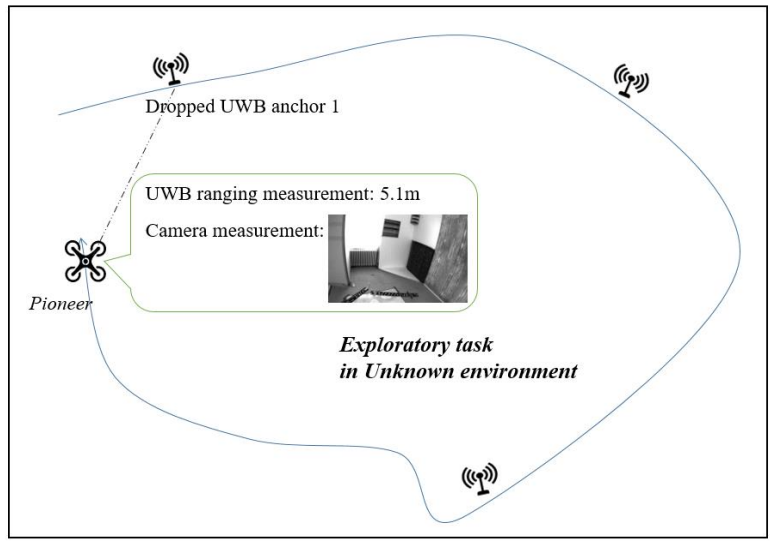

Fig.1. Exploratory task: a pioneering vehicle carrying UWB anchors explores an unknown environment. During exploration, the anchors will be deployed. A visual-UWB SLAM algorithm is performed for visual mapping and anchor positioning while tracking the motion of the vehicle.

Then, our system reuses the built map for navigation tasks as shown in Fig.2. Different from the general visual localization method that reuses the map by finding the correspondences of features with the map, our localization method can reuse the map by ranging to the UWB anchors whose positions were previously mapped. In contrast to a pioneer that demands rich onboard resources, vehicles with only a UWB tag and poor computation resource are capable of self-localization precisely. Aided with the comprehensive representation of the environment in the visual map, our system allows the navigation tasks such as path planning and collision avoidance. Note that the localization of a pioneer-like vehicle by reusing the map is also provided and the map can be reused online (during the exploratory task) or off-line (after the exploratory task).

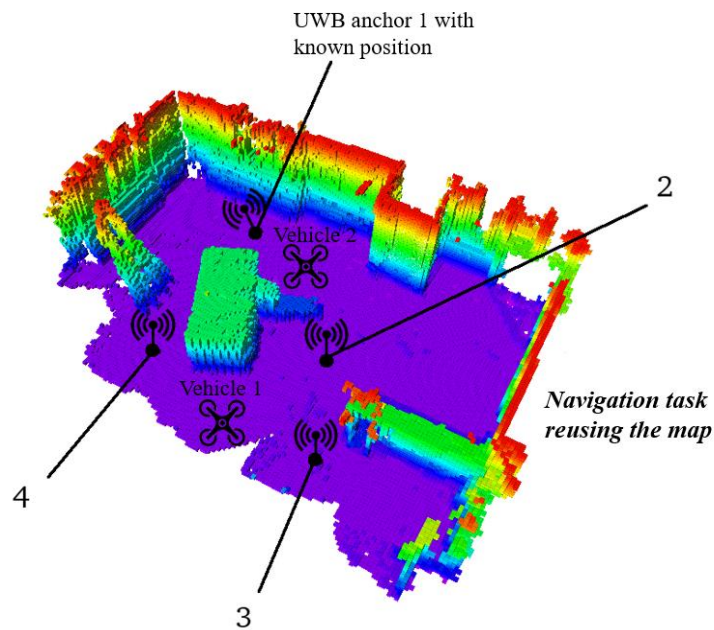

Fig.2. Navigation task: once the map is available, vehicles carrying only a UWB tag can determine their own position.

The remainder of this paper is structured as follows. First, we provide a summary of the relevant sensor preliminaries. Then we detail the exploratory task by introducing the problem formulation and the optimization algorithm on Lie-Manifold. Next, we provide a guidance on how to reuse the map in a navigation task. Finally, tests on public datasets are conducted to evaluate the performance of our visual-UWB navigation system.

## PRELIMINARY

**Notation and definition**

Notation employed in the work is defined: the three-dimensional point **p** expressed in a reference frame $A$ is written as $\mathbf{p}_A$, the vector expressed in a reference frame $A$ is written as $\mathbf{t}_{pq,A}$ with **p** and **q** as start and end points respectively, the rotation matrix $\mathbf{R}_{AB}$ represents rotation from frame $B$ to frame $A$. We define the camera frame as $C$, the world reference frame as $W$, and the body frame as $B$ which coincides with the UWB frame. For simple expression, we omit the subscript $W$ for quantities expressed in world reference frame.

Consider vehicle motion in three-dimensional space, the pose of the vehicle can be denoted as an invertible $4 \times 4$ transformation matrix **T** in special Euclidean group SE(3). It turns out that SE(3) is a Lie group and a manifold with structure $SO(3) \times \mathbb{R}^3$. SO(3) is the group of 3D rotation matrices that form a smooth manifold [10]. The pose can be described by means of a translation vector **t** and a rotation matrix **R**:

$$\mathbf{T} = [\mathbf{R} \mid \mathbf{t}] = \begin{bmatrix} \mathbf{R} & \mathbf{t} \\ \mathbf{0}^T & 1 \end{bmatrix} \in SE(3), \quad \mathbf{R} \in SO(3), \quad \mathbf{t} \in \mathbb{R}^3 \tag{1}$$

$\mathbf{T}_{AB}$ represents the transformation from frame B to frame A.

**Visual-UWB preliminaries**

In our system, the UWB transceiver module mounted on the vehicle is designated as a tag, ranging with a set of anchors in a round-robin schedule. A very precise clock is implemented in the UWB modules, thus the accurate transmission and reception time of the UWB signal was recorded. To reduce the error due to clock shift and frequency shift, symmetric double-sided two way ranging (TWR) method can be used to obtain the TOF measurements [11]. The ranging measurement between the tag and a corresponding anchor $k$ can be modeled as:

$$D_k = c * t_f = \|\mathbf{p}^k - \mathbf{p}^b\| + n_k \tag{2}$$

where $\mathbf{p}^b$ is the position of the UWB tag (body frame center) and $\mathbf{p}^k$ is the position of the anchor k, both expressed in world reference frame with subscript $W$ omitted for simple expression. And $n_k$ is the measurement noise assumed to be a zero mean Gaussian noise, $n_k \sim \mathcal{N}(0, {\sigma_{uwb}^k}^2)$, $c$ the transmission speed of radio wave assumed to be the same as light speed and $t_f$ the measured TOF.

A monocular camera is used with perspective projection model $h(\cdot)$, which project 3D points $\mathbf{p}_C = [X_C \ Y_C \ Z_C]^T \in \mathbb{R}^3$ in camera frame $C$ into 2D points on the image plane $\mathbf{z} \in \Omega \subset \mathbb{R}^2$:

$$\mathbf{z} = h(\mathbf{p}_C) \tag{3}$$

Our system considers a conventional pinhole-camera projection model:

$$\mathbf{z} = h(\mathbf{p}_C) = \begin{bmatrix} \dfrac{f_u X_C}{Z_C} + c_u \\ \dfrac{f_v Y_C}{Z_C} + c_v \end{bmatrix} \tag{4}$$

where $[f_u \ f_v]$ is the focal length and $[c_u \ c_v]$ is the principal point [12].

The monocular camera and the UWB tag are both rigidly attached to the pioneering vehicle and the camera-UWB extrinsic matrix $\mathbf{T}_{CB} = [\mathbf{R}_{CB} \mid \mathbf{t}_{CB}]$ is pre-calibrated and known, typically $\mathbf{R}_{CB}$ equals identity matrix. The UWB ranging measurements and the monocular camera images are fed into our system with aligned timestamps.

## EXPLORATORY TASK BY VISUAL-UWB SLAM

**Problem Formulation**

The aim of our proposed visual-UWB SLAM is simultaneous tracking of the vehicle's motion and mapping of the anchors and landmarks in the environment. Let us denote the variables to be estimated as $\mathbf{X}$ which comprise the vehicle pose $\mathbf{X}_v$, the position of the deployed anchors $\mathbf{p}^k$ and landmarks $\mathbf{p}^L$. To save computational memory, we use a unit quaternion to present the orientation of the vehicle. Thus, the pose variable can be written as $\mathbf{X}_v := [\mathbf{q}_{BW}^T, \mathbf{t}_{BW}^T]^T$. To formulate the tightly-coupled visual-UWB SLAM, we incorporate the ranging error term $\mathbf{e}_{uwb}^{i,k}$ with the reprojection error $\mathbf{e}_{rp}^{i,j}$ into one cost function $C(\mathbf{X})$ as one joint optimization:

$$C(\mathbf{X}) = \sum_{i,j} \rho\left(\mathbf{e}_{rp}^{i,j}{}^T \Sigma_{rp}^{i,j}{}^{-1} \mathbf{e}_{rp}^{i,j}\right) + \sum_{i,k} \rho\left(\mathbf{e}_{uwb}^{i,k}{}^T \Sigma_{uwb}^{k}{}^{-1} \mathbf{e}_{uwb}^{i,k}\right) \tag{5}$$

where $i$ is the index of the monocular camera frame, $j$ the index of landmark and $k$ the index of UWB anchor. And $\rho(\cdot)$ is the Huber loss function, $\Sigma_{rp}^{i,j} = \sigma_{rp}^{i,j}{}^2 \mathbf{I}_{2\times 2}$ is the covariance matrix of the respective landmark measurement, $\Sigma_{uwb}^{k} = \sigma_{uwb}^{k}{}^2$ the covariance of the UWB ranging measurement. The variables are estimated by solving the following least squares minimization problem:

$$\mathbf{X}^* = \underset{\mathbf{X}}{\operatorname{argmin}}\, C(\mathbf{X}) \tag{6}$$

Consider the landmark (index $j$) $\mathbf{p}^{L,j}$ that is observed in the frame (index $i$), the reprojection error term is:

$$\mathbf{e}_{rp}^{i,j} = \mathbf{z}^{i,j} - h(\mathbf{p}_C^{i,j}) \tag{7}$$
$$\mathbf{p}_C^{i,j} = \mathbf{R}_{CB}(\mathbf{R}_{BW}^i \mathbf{p}^{L,j} + \mathbf{t}_{BW}) + \mathbf{t}_{CB} \tag{8}$$

where $\mathbf{z}^{i,j}$ is the measured keypoint in the image plane.

The error term of the UWB ranging measurement is designed based on (2):

$$\mathbf{e}_{uwb}^{i,k} = D_k - \|\mathbf{p}^k - \mathbf{p}^{b,i}\| = D_k - \|\mathbf{p}^k + \mathbf{R}_{BW}^i{}^T \mathbf{t}_W^i\| \tag{9}$$

**Optimization on Lie-manifold**

Generally, if a good initial guess of the variable is obtained, the least squares minimization problem in Euclidean space can be solved by the popular Gauss-Newton or Levenberg-Marquardt algorithms [13]. It works by approximating the cost function to a quadratic form and reduces to solve the linear equations repeatedly. The solution of this local approximation is then used as an increment to update the initial guess. However, the pose variable in SE(3) spans over a non-Euclidean space (belongs to a manifold), thus the general Gauss-Newton or Levenberg-Marquardt algorithm cannot be directly applied [14]. A common approach is to apply the optimization problem on manifold whose local tangent space in current estimate behaves as a Euclidean space. And we define the operator $\boxplus$ that maps the disturbance $\Delta x$ in tangent space to a disturbance on the manifold around the estimate $\overline{\mathbf{X}}$, $\mathbf{X}^* = \overline{\mathbf{X}} \boxplus \Delta x$. A detailed introduction of SE(3) Lie-manifold can be found in [10,15]. By replacing the operator $+$ with $\boxplus$, the framework of conventional optimization algorithm can be applied to the optimization problem on manifold.

For practical implementation in our system, we only consider the SO(3) manifold, as the translation part of the vehicle pose clearly forms a Euclidean space. For pose variable $\mathbf{X}_v$, we represent the local disturbance $\Delta \mathbf{x}_v$ in tangent space as 6D vector $\Delta \mathbf{x}_v := [\Delta \mathbf{t}^T, \Delta \boldsymbol{\omega}^T]^T \in \mathbb{R}^6$, where $\Delta \mathbf{t} \in \mathbb{R}^3$ is the 3D translation and $\Delta \boldsymbol{\omega} \in \mathbb{R}^3$ is the rotation. The Jacobians of the reprojection error with respect to the pose disturbance $\Delta \mathbf{x}_v$ follow from [10]:

$$\mathbf{J}_{\mathbf{X}_v}^{rp} = \left.\frac{\partial \mathbf{e}_{rp}^{i,j}(\mathbf{X}_v \boxplus \Delta \mathbf{x}_v)}{\partial \Delta \mathbf{x}_v}\right|_{\Delta \mathbf{x}_v = 0}$$
$$= -\begin{bmatrix} \frac{f_u}{z} & 0 & -\frac{f_u x}{z^2} \\ 0 & \frac{f_v}{z} & -\frac{f_v y}{z^2} \end{bmatrix} \mathbf{R}_{CB}\begin{bmatrix} -\mathbf{p}_B^{L\wedge} & \mathbf{I}_{3\times 3} \end{bmatrix} \tag{10}$$

where $\mathbf{p}_B^L = \begin{bmatrix} x & y & z \end{bmatrix}^T$ is the coordinate of landmark expressed in the body frame, $(\cdot)^\wedge$ denotes the skew-symmetric cross-product matrix determined by a vector in $\mathbb{R}^3$. The Jacobians of the reprojection error with respect to the landmark position disturbance is given as:

$$\mathbf{J}_{\mathbf{p}^L}^{rp} = - \begin{bmatrix} \dfrac{f_u}{z} & 0 & -\dfrac{f_u x}{z^2} \\ 0 & \dfrac{f_v}{z} & -\dfrac{f_v y}{z^2} \end{bmatrix} \mathbf{R}_{CB} \mathbf{R}_{BW} \quad (11)$$

The Jacobians of the UWB ranging error with respect to the pose disturbance $\Delta \mathbf{x}_v$ is straightforward to obtain:

$$\mathbf{J}_{\mathbf{X}_v}^{uwb} = \frac{1}{\|\mathbf{p}^k - \mathbf{p}^b\|} \begin{bmatrix} -\left(\mathbf{R}_{BW}(\mathbf{p}^k - \mathbf{p}^b)\right)^T & \mathbf{0}_{3\times 3} \end{bmatrix} \quad (12)$$

Finally, the Jacobians with respect to the anchor position disturbance is given as:

$$\mathbf{J}_{\mathbf{p}^k}^{uwb} = -\frac{(\mathbf{p}^k - \mathbf{p}^b)^T}{\|\mathbf{p}^k - \mathbf{p}^b\|} \quad (13)$$

By defining the above Jacobians with respect to the corresponding variables, we solve the least squares minimization problem utilizing Gauss-Newton algorithm implemented in General Graph Optimization (g2o) toolbox [16].

**Visual-UWB SLAM**
We build our visual-UWB SLAM system based on the frontend framework of ORB-SLAM [4], which is one of the state-of-art monocular visual SLAM systems. Note that our navigation system can also incorporate with other visual SLAM systems. Inherited from ORB-SLAM, the processing pipeline of our visual-UWB SLAM consists of three parallel thread: tracking, local mapping, and loop closing.

In tracking thread, we aim to track the real-time pose of the vehicle. The initial pose of the vehicle is predicted by a constant-velocity motion model. We then project the landmarks in the local map into the current captured monocular image and match the landmarks with the image keypoints. Typically, camera images and UWB ranging measurements are acquired at a different frequency, e.g. at 25Hz and 50Hz respectively. For the current image time instant *T*, we associate a UWB ranging measurement with this image whose timestamp is closest to *T*. The vehicle pose is then optimized by minimizing the reprojection errors of matched features and the associated UWB ranging error. This is a motion-only bundle adjustment as we only optimize the pose of the vehicle. The cost function is the same as (5) and all the landmarks and UWB anchors are fixed, only the pose variable $\mathbf{X}_v$ is optimized. Note that the tracking process is also the core process in localization task, as the map is fixed and not updated.

Once a new keyframe is created in tracking thread and inserted into the map, the visual-UWB local mapping thread performs a local bundle adjustment. It optimizes the positions of all deployed UWB anchors, the pose of the current keyframe, all other keyframes connected to current keyframe (i.e. share observations of landmarks with current keyframe and the number of co-visible landmarks should be at least e.g. 15), and all the co-visible landmarks using equation (5). The initial position of the anchor is the position of the vehicle where are deployed. Keyframes that observe these landmarks but are not connected to current keyframe also contribute to this optimization with their poses remain fixed. The position of the very first deployed anchor is also set as fixed in optimization, as it's set to be the origin of the world reference frame during the initialization process. Fig.3. shows this process using a graph representation.

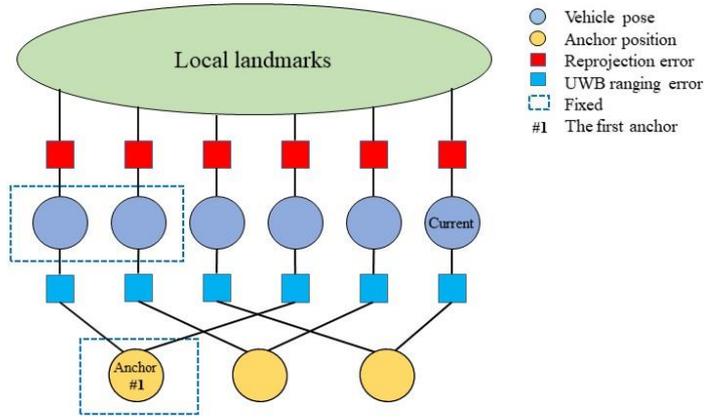

Fig.3. Graph representation of local bundle adjustment. The variables to be estimated are depicted by the round boxes and are connected by constraints depicted by the square boxes. The constraints consist of the reprojection errors and the UWB ranging errors. A local bundle adjustment optimizes the poses of the local keyframes, all the local co-visible landmarks and the deployed anchors in the environment. The pose of the keyframe that is not connected to current keyframe but shares observations of these local landmarks is fixed during optimization.

The loop closing thread detects whether the vehicle returned to an already explored area to reduce the accumulated drift. For a newly created keyframe, loop detection is performed by querying the past keyframes database. Once matched to a past keyframe, temporal and geometrical consistency was validated. Finally, a pose-graph optimization is performed to reduce the accumulated error in the past trajectory. Since UWB ranging measurements renders the scale factor observable, in contrast to ORB-SLAM, the pose-graph optimization only considers 6DoF instead of 7DoF. UWB ranging measurements are not used for pose-graph optimization, the anchor positions are not optimized. Afterward, a full bundle adjustment is applied to optimize all variables and improve accuracy.

**Initialization**
Based on the assumption that the scale is arbitrary, and the initial pose coinciding with the origin of world reference frame, the visual SLAM system can bootstrap itself by computing the ego-motion of first two keyframes. However, to directly fuse UWB measurements with visual measurements, we need to recover the scale. This is the aim of our initialization process.

Our system first performs the visual SLAM initialization process. Once visual SLAM is initialized, a first-ever anchor is dropped and deployed. The position $\mathbf{p}^1$ of the first anchor is assumed to be at precisely the same position of the vehicle body, i.e. the position of the first anchor and the position of the vehicle body both coincide with the origin of world reference frame, thus we have $\mathbf{p}^1 = [0\ \ 0\ \ 0]^T$. Note that world reference frame is defined by coinciding with the initial pose of the vehicle. Then we run the visual SLAM to retrieve at least 10 keyframes and separately obtain associated UWB ranging measurements to render the scale observable. Consider the output pose $\mathbf{T}_{CW} = [\mathbf{R}_{CW} \mid \mathbf{t}_{CW}]$ of keyframe from visual SLAM, we can obtain the body center $\mathbf{p}^b$ in world reference frame as:

$$\mathbf{p}^b = s\mathbf{p}^c + \mathbf{R}_{CW}^T \mathbf{t}_{CB} \tag{14}$$

where $\mathbf{p}^c = -\mathbf{R}_{CW}^T \mathbf{t}_{CW}$ is the camera center expressed in world reference frame. And we have the associated UWB ranging measurements $D_1$ from the first anchor. Substituting (14) into (2) and rearranging we have:

$$D_1 = \|s\mathbf{p}^c + \mathbf{R}_{CW}^T \mathbf{t}_{CB}\| \tag{15}$$

The scale $s$ is then determined by minimizing:

$$s^* = \underset{s}{\mathrm{argmin}} \sum_i D_1^i - \|s\mathbf{p}^{c,i} + \mathbf{R}_{CW}^{i\ T} \mathbf{t}_{CB}\| \tag{16}$$

where $i$ is the index of keyframes used for initialization.

After initialization completion, the initially built map and the poses of the past keyframes will be scaled by the scale factor.

**Anchor constellation**

In this part, we introduce the strategy on how to deploy the UWB anchors. For localization task only using the UWB ranging measurements, the localization accuracy depends on the constellation of the UWB anchors. The anchors must be distinctly separately in 3D space, this is a well-known problem from Global Navigation Satellite System (GNSS) [17]. Generally, a larger number of anchors would decrease the Dilution of Precision (DOP) values, yield better performance. On the other hand, the cost of a practical application implementation would increase. As a trade-off, during the exploratory task, an anchor is deployed if the vehicle is far away (e.g. 20 meters away) from the previously deployed anchors. Especially, the first anchor is dropped when the visual SLAM initialization is completed, then the second anchor is dropped after the visual-UWB SLAM initialization. Afterward, an incremental constellation of anchors is constructed by dropping anchors during exploration. Alternatively, the stream of camera images and the current building map can be transferred to a remote expert who is experienced in the construction of constellation, thus manually decide when to deploy an anchor.

## LOCALIZATION TASK BY REUSING THE MAP

Once an exploratory task is completed, the comprehensive map of the environment is obtained. Navigation can be done against the map. Depending on the sensors attached to the vehicle that conducts a navigation task, different localization method is used: UWB-only localization and visual-UWB localization method.

For UWB-only localization method, the position of the vehicle can be determined from UWB ranging measurements using the conventional triangulation method.

Visual-UWB localization method is used for complex navigation tasks that require 3D incremental pose estimation. Both the monocular camera and the UWB tag are equipped with the vehicle. The localization is performed like the visual-UWB SLAM algorithm precluding the mapping and loop closing thread as we detailed in the previous section, as the map is available and not updated.

Compared with the exploratory task, the localization task requires little computational resources as the map is readily available, thus enabling the localization of small robots with limited payload capability. Different vehicles in terms of equipped sensors and payload capability in a heterogeneous team can concurrently perform localization task by reusing the same map. Hence, our system is extendable to groups of vehicles.

## EXPERIMENTS

We evaluate the performance of our proposed visual-UWB navigation system using the EuRoC MAV Datasets [18]. The datasets are collected onboard a MAV, which contains the stereo images (Aptina MT9V034 global shutter, WVGA monochrome, 20Hz), synchronized IMU measurements and ground truth vehicle state (Vicon motion capture system, 100Hz). We test our system on the dataset sequence collected in an industrial environment as shown in Fig.4. We only use the images from the left camera and use the ground truth to simulate the sequence of UWB ranging measurements. To simulate the UWB anchor deployment during the exploratory task, we manually record the timestamp when the visual ORB-SLAM completes its initialization process, then set the ground truth position of the MAV at the recorded time to be the first deployed anchor position. Five anchors are deployed one by one in a random manner, therefore showing the versatility of our proposed system for different anchor constellation construction strategy. In detail, the anchors are deployed approximately every ten seconds along the ground truth trajectory. The UWB ranging measurements are simulated at the frequency of 100Hz, and the standard deviation of the additive Gaussian noise is 0.01m. The UWB tag only performs TWR with the available anchors (deployed ones) in the environment by a round-robin scheme, i.e. if 2 anchors are deployed, then the simulated UWB ranging measurements from anchor 1 is 50 Hz.

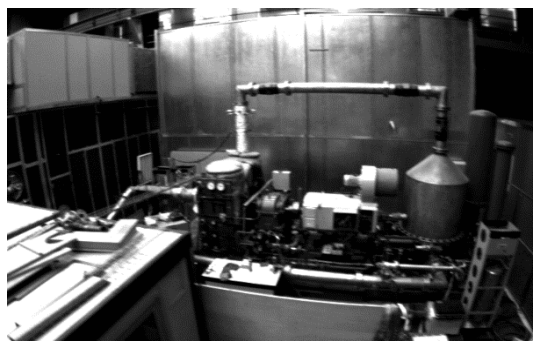

Fig.4. A sample image of the EuRoC datasets collected in an industrial environment.

We evaluate our visual-UWB SLAM mainly in term of the accuracy of the estimated position of UWB anchors and the recovered MAV trajectory. The initialization process is performed when 5 keyframes are created. For better comparison, the estimated trajectory is aligned with the ground truth in 6DoF. The results plotting and alignment methods are provided by the Python package *evo [19]*. In order to evaluate the initialization process, we also align the estimated trajectory with ground truth in 7DoF to measure the ideal scale factor. This scale factor can be considered as the residual scale error. Our system performs on dataset sequence *MH_03_medium* in real-time, the estimated MAV trajectory and the ground truth are shown in Fig.5. The absolute translation error of the trajectory versus time is shown in Fig.6. The error is bounded to [0.034, 0.120]m and its mean is 0.036m.

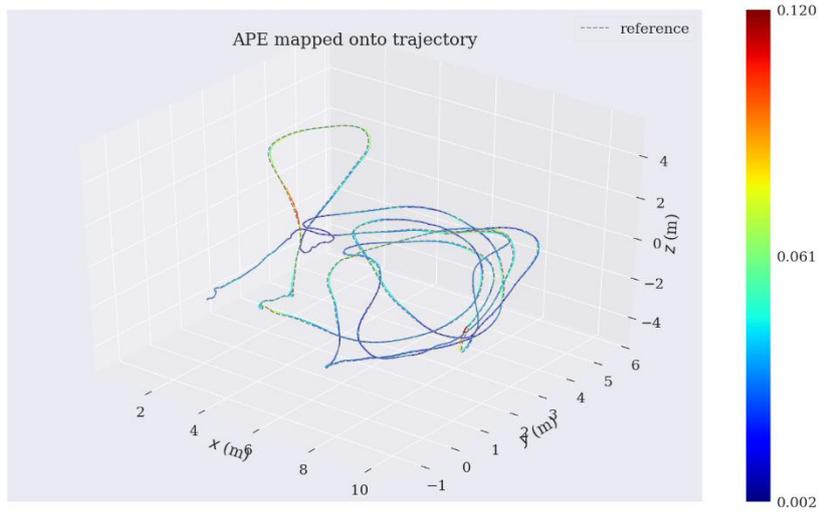

Fig.5. The comparison of estimated trajectory and ground truth (6 DoF aligned). The estimated trajectory is plotted using a colored line with red denoting large error and blue denoting small error, and the reference ground truth is depicted by a dotted line.

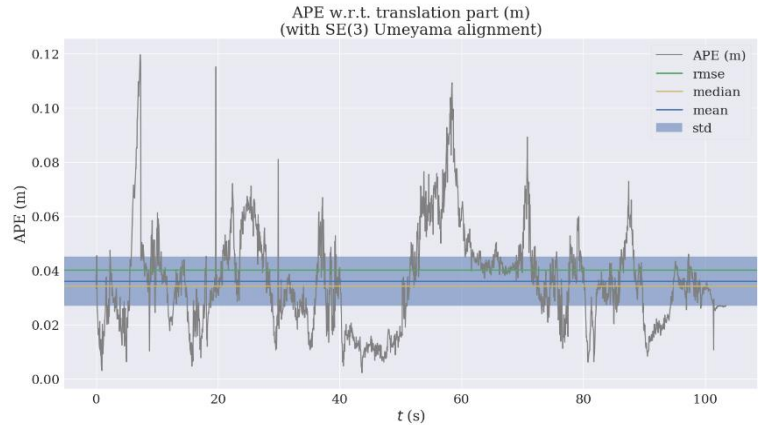

Fig.6. Translation error of MAV trajectory versus time.

The optimal SE(3) matrix aligning the estimated trajectory with the ground truth is applied to the estimated anchor positions, thus we can evaluate the accuracy of the anchor positions against the ground truth. Table. 1. shows the final estimated anchor positions and the absolute error with respect to ground truth. The average error in meters is 0.025.

**Table. 1.** The estimated anchor positions and their error with ground truth.

| Anchors | Estimate (m) | Ground truth (m) | Error (m) |
| --- | --- | --- | --- |
| 1 | (3.635, 0.632, 0.854) | (3.643, 0.600, 0.864) | 0.034 |
| 2 | (6.742, 3.458, 0.523) | (6.726, 3.4490, 0.506) | 0.025 |
| 3 | (8.411, 3.653, -0.247) | (8.407, 3.660, -0.228) | 0.021 |
| 4 | (10.422, 0.887, -0.760) | (10.409, 0.889, -0.779) | 0.023 |
| 5 | (8.919, -1.750, 0.047) | (8.905, -1.731, 0.041) | 0.024 |

The scale factor error is obtained as 0.28% using a 7DoF alignment.

As we can see, our system is capable of online determination of the settled UWB anchors during the exploratory task, achieving a typical precision of 2.5cm. And the motion of the MAV is estimated with a metric scale, with a scale error typically below 0.3%. The precision of the recovered MAV trajectory is 3.6cm for 300m$^2$ industrial environments.

To evaluate the localization task by reusing the map, we replay the dataset in a localization mode. Our system performs relocalization process for the first incoming image and continues tracking the motion of the MAV using the previous map. Here we use both the UWB ranging measurements and the visual measurements, the UWB-only localization method is omitted for evaluation as it's simple and straightforward. The absolute translation error is 3.5cm, showing that our system is able to reuse the map built in the exploratory task and the localization method suffers no drift error.

**CONCLUSION**

In this paper, we propose a visual-UWB navigation system for unknown environments without UWB anchors as a prior. Our system first performs an exploratory task when the vehicle first enters an unknown environment. During the task, the vehicle drops and settles some UWB anchors and online estimate the position of these anchors. Furthermore, our visual-UWB SLAM recovers the pose of the vehicle with metric scale and build a visual map of the environment. Then the map is shared with the resource-limited vehicles carrying out a localization task which require little computational capability. We have experimentally evaluated our system on the benchmark datasets. The results show that during an exploratory task our system is able to estimate the position of the anchors and recover the motion of the vehicle with a metric scale, achieving a typical centimeter precision. And by reusing the map for localization tasks, the localization method also achieves centimeter precision.